\documentclass[11pt,a4paper]{article}
\usepackage[hyperref]{emnlp2018}
\usepackage{times}
\usepackage{latexsym}
\usepackage{amsmath,amssymb}
\usepackage{graphicx}
\usepackage{bm}
\usepackage{textcomp}
\usepackage{multirow}

\usepackage{url}
\DeclareMathOperator{\E}{\mathbb{E}}

\graphicspath{{figures/}}

\aclfinalcopy 
\def\aclpaperid{585} 

\def\figref#1{Fig.~\ref{#1}}

\def\tabref#1{Table~\ref{#1}}
\def\eqnref#1{Eqn.~\ref{#1}}

\title{Closed-Book Training to Improve Summarization Encoder Memory}

\author{Yichen Jiang \and Mohit Bansal \\
  UNC Chapel Hill \\
  {\tt \{yichenj, mbansal\}@cs.unc.edu} \\
 }

\date{}

\begin{document}
\maketitle

\begin{abstract}

A good neural sequence-to-sequence summarization model should have a strong encoder that can distill and memorize the important information from long input texts so that the decoder can generate salient summaries based on the encoder's memory. In this paper, we aim to improve the memorization capabilities of the encoder of a pointer-generator model by adding an additional `closed-book' decoder without attention and pointer mechanisms. Such a decoder forces the encoder to be more selective in the information encoded in its memory state because the decoder can't rely on the extra information provided by the attention and possibly copy modules, and hence improves the entire model. 
On the \emph{CNN/Daily Mail} dataset, our 2-decoder model outperforms the baseline significantly in terms of ROUGE and METEOR metrics, for both cross-entropy and reinforced setups (and on human evaluation). 
Moreover, our model also achieves higher scores in a test-only DUC-2002 generalizability setup.
We further present a memory ability test, two saliency metrics, as well as several sanity-check ablations (based on fixed-encoder, gradient-flow cut, and model capacity) to prove that the encoder of our 2-decoder model does in fact learn stronger memory representations than the baseline encoder.
\end{abstract}
\section{Introduction}
 
\begin{figure}[t]
\centering
\includegraphics[width=0.45\textwidth]{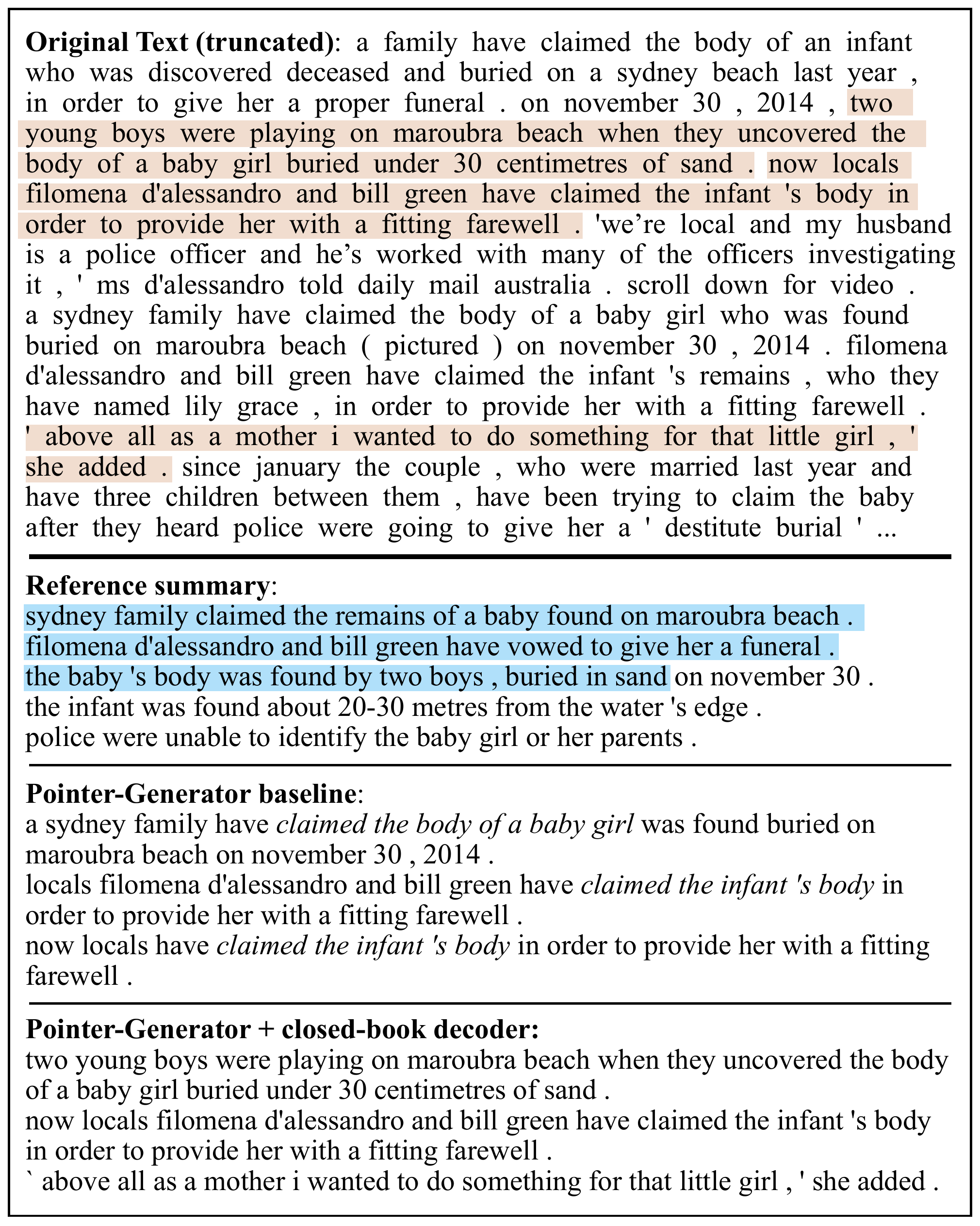}
\vspace{-6pt} 
\caption{Baseline model repeats itself twice (italic), and fails to find all salient information (highlighted in red in the original text) from the source text that is covered by our 2-decoder model.
The summary generated by our 2-decoder model also recovers most of the information mentioned in the reference summary (highlighted in blue in the reference summary).\label{fig:example-infant-body-small} }
\end{figure}

Text summarization is the task of condensing a long passage to a shorter version that only covers the most salient information from the original text. Extractive summarization models \cite{Jing:00,Knight:02,Clarke:08,Filippova:15} directly pick words, phrases, and sentences from the source text to form a summary, while an abstractive model generates (samples) words from a fixed-size vocabulary instead of copying from text directly. 

The last few years have seen significant progress on both extractive and abstractive approaches, of which a large number of studies are fueled by neural sequence-to-sequence models~\cite{Sutskever:14}. One popular formulation of such models is an RNN/LSTM encoder that encodes the source passage to a fixed-size memory-state vector, and another RNN/LSTM decoder that generates the summary from this memory state. This paradigm is enhanced by attention mechanism~\cite{Bahdanau:14} and pointer network~\cite{Vinyals:15}, such that the decoder can refer to (and weigh) all the encoding steps' hidden states or directly copy words from the source text, instead of relying solely on encoder's final memory state for all information about the source passage.
Recent studies \cite{Rush:15,Nallapati:16,Chopra:16,Zeng:16,Gu:16,Gulcehre:16,See:17} have demonstrated success with such seq-attention-seq and pointer models in summarization tasks.

While the advantage of attention and pointer models compared to vanilla sequence-to-sequence models in summarization is well supported by previous studies, these models still struggle to find the most salient information in the source text when generating summaries. This is because summarization, being different from other text-to-text generation tasks (where there is an almost one-to-one correspondence between input and output words, e.g., machine translation), requires the sequence-attention-sequence model to additionally decide where to attend and where to ignore, thus demanding a strong encoder that can determine the importance of different words, phrases, and sentences and flexibly encode salient information in its memory state.
To this end, we propose a novel 2-decoder architecture by adding another `closed book' decoder without attention layer to a popular pointer-generator baseline, such that the `closed book' decoder and pointer decoder share an encoder. 
We argue that this additional `closed book' decoder encourages the encoder to be better at memorizing salient information from the source passage, and hence strengthen the entire model.
We provide both intuition and evidence for this argument in the following paragraphs.

Consider the following case. Two students are learning to do summarization from scratch. During training, both students can first scan through the passage once (encoder's pass). Then student A is allowed to constantly look back (attention) at the passage when writing the summary (similar to a pointer-generator model), while student B has to occasionally write the summary without looking back (similar to our 2-decoder model with a non-attention/copy decoder). During the final test, both students can look at the passage while writing summaries. We argue that student B will write more salient summaries in the test because s/he learns to better distill and memorize important information in the first scan/pass by not looking back at the passage in training.

\label{sec 1}
In terms of back-propagation intuition, during the training of a seq-attention-seq model (e.g.,~\newcite{See:17}), most gradients are back-propagated from the decoder to the encoder's hidden states through the attention layer. This encourages the encoder to correctly encode salient words at the corresponding encoding steps, but does make sure that this information is not forgotten (overwritten in the memory state) by the encoder afterward. 
However, for a plain LSTM (closed-book) decoder without attention, its generated gradient flow is back-propagated to the encoder through the memory state, which is the only connection between itself and the encoder, and this, therefore, encourages the encoder to encode only the salient, important information in its memory state.
Hence, to achieve this desired effect, we jointly train the two decoders, which share one encoder, by optimizing the weighted sum of their losses. 
This approximates the training routine of student B because \emph{the sole encoder has to perform well for both decoders}.
During inference, we only employ the pointer decoder due to its copying advantage over the closed-book decoder, similar to the situation of student B being able to refer back to the passage during the test for best performance (but is still trained hard to do well in both situations).
\figref{fig:example-infant-body-small} shows an example of our 2-decoder summarizer generating a summary that covers the original passage with more saliency than the baseline model.

Empirically, we test our 2-decoder architecture on the~\emph{CNN/Daily Mail} dataset~\cite{Hermann:15,Nallapati:16}, and our model surpasses the strong pointer-generator baseline significantly on both ROUGE~\cite{Lin:04} and METEOR~\cite{Denkowski:14} metrics, as well as based on  human evaluation. This holds true both for a cross-entropy baseline as well as a stronger, policy-gradient based reinforcement learning setup~\cite{Williams:92}. 
Moreover, our 2-decoder models (both cross-entropy and reinforced) also achieve reasonable improvements on a test-only generalizability/transfer setup on the DUC-2002  dataset.

We further present a series of numeric and qualitative analysis to understand whether the improvements in these automatic metric scores are in fact due to the enhanced memory and saliency strengths of our encoder. 
First, by evaluating the representation power of the encoder's final memory state after reading long passages (w.r.t. the memory state after reading ground-truth summaries) via a cosine-similarity test, we prove that our 2-decoder model indeed has a stronger encoder with better memory ability.
Next, we conduct three sets of ablation studies based on fixed-encoder, gradient-flow cut, and model capacity to show that the stronger encoder is the reason behind the significant improvements in ROUGE and METEOR scores.
Finally, we show that summaries generated by our 2-decoder model are qualitatively better than baseline summaries as the former achieved higher scores on two saliency metrics (based on cloze-Q\&A blanks and a keyword classifier) than the baseline summaries, while maintaining similar length and better avoiding repetitions. This directly demonstrates our 2-decoder model's enhanced ability to memorize and recover important information from the input document, which is our main contribution in this paper.

\section{Related Work}

\noindent\textbf{Extractive and Abstractive Summarization}:
Early models for automatic text summarization were usually extractive \cite{Jing:00,Knight:02,Clarke:08,Filippova:15}.
For abstractive summarization, different early non-neural approaches were applied, based on graphs~\cite{Giannakopoulos:09,Ganesan:10}, discourse trees~\cite{Gerani:14}, syntactic parse trees~\cite{Cheung:14,Wang:16}, and a combination of linguistic compression and topic detection~\cite{Zajic:04}.
Recent neural-network models have tackled abstractive summarization using methods such as hierarchical encoders and attention, coverage, and distraction~\cite{Rush:15,Chopra:16,Nallapati:16,Chen2016DistractionBasedNN,Takase:16} as well as various initial large-scale, short-length summarization datasets like DUC-2004 and Gigaword.
\newcite{Nallapati:16} adapted the \emph{CNN/Daily Mail} \cite{Hermann:15} dataset for long-text summarization, and provided an abstractive baseline using attentional sequence-to-sequence model. 

\noindent\textbf{Pointer Network for Summarization}:
Pointer networks~\cite{Vinyals:15} are useful for summarization models because summaries often need to copy/contain a large number of words that have appeared in the source text. This provides the advantages of both \emph{extractive} and \emph{abstractive} approaches, and usually includes a gating function to model the distribution for the extended vocabulary including the pre-set vocabulary and words from the source text~\cite{Zeng:16,Nallapati:16,Gu:16,Gulcehre:16,Miao:16}.
\newcite{See:17} used a soft gate to control model's behavior of copying versus generating. 
They further applied coverage mechanism and achieved the state-of-the-art results on \emph{CNN/Daily Mail} dataset. 

\noindent\textbf{Memory Enhancement}:
Some recent works \cite{Wang:16memory,Xiong:17,Gu:16b} have studied enhancing the memory capacity of sequence-to-sequence models. They studied this problem in Neural Machine Translation by keeping an external memory state analogous to “data” in the Von
Neumann architecture, while the “instructions” are represented by the sequence-to-sequence model. Our work is novel in that we aim to improve the internal long-term memory of the encoder LSTM by adding a closed-book decoder that has no attention layer, yielding a more efficient internal memory that encodes only important information from the source text, which is crucial for the task of long-document summarization.

\noindent\textbf{Reinforcement Learning}: Teacher forcing style maximum likelihood training suffers from exposure bias \cite{Bengio:15}, so recent works instead apply reinforcement learning style policy gradient algorithms (REINFORCE~\cite{Williams:92}) to directly optimize on metric scores~\cite{Henss:15,Paulus:17}. 
Reinforced models that employ this method have achieved good results in a number of tasks including image captioning \cite{Liu:16,Ranzato:16}, machine translation \cite{Bahdanau:16,Norouzi:16}, and text summarization \cite{Ranzato:16,Paulus:17}.

\section{Models}

\begin{figure*}[t]
\centering
\includegraphics[width=0.98\textwidth]{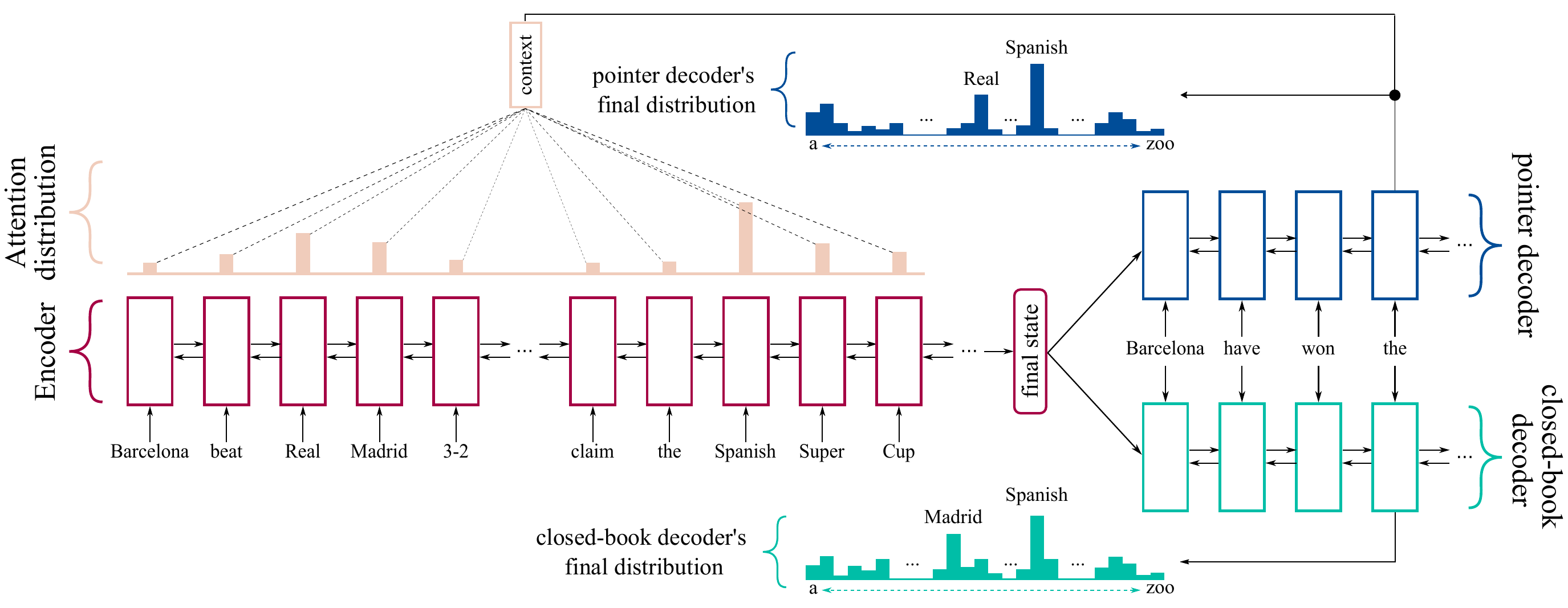}
\vspace{-10pt} 
\caption{Our 2-decoder summarization model with a pointer decoder and a closed-book decoder, both sharing a single encoder (this is during training; next, at inference time, we only employ the memory-enhanced encoder and the pointer decoder). \label{fig:model} \vspace{-4pt}}
\end{figure*}

\subsection{Pointer-Generator Baseline}
The pointer-generator network proposed in \newcite{See:17} can be seen as a hybrid of \emph{extractive} and \emph{abstractive} summarization models. At each decoding step, the model can either sample a word from its vocabulary, or copy a word directly from the source passage. 
This is enabled by the attention mechanism \cite{Bahdanau:14}, which includes a distribution $a_{i}$ over all encoding steps, and a context vector $c_t$ that is the weighted sum of encoder's hidden states. 
The attention mechanism is modeled as:
\begin{equation} \label{eq:att}
\begin{split}
e_{i}^{t} & = v^T \mathrm{tanh}(W_h h_i + W_s s_t + b_{attn}) \\
a_{i}^t & = \mathrm{softmax}(e_{i}^t); \;\;\; c_t = \sum_{i}a_{i}^{t} h_{i}
\end{split}
\vspace{-10pt}
\end{equation}
where $v$, $W_h$, $W_s$, and $b_{attn}$ are learnable parameters. $h_i$ is encoder's hidden state at $i^{th}$ encoding step, and $s_t$ is decoder's hidden state at $t^{th}$ decoding step. 
The distribution $a_{i}^{t}$ can be seen as the amount of attention at decode step $t$ towards the $i^{th}$ encoder state. Therefore, the context vector $c_t$ is the sum of the encoder's hidden states weighted by attention distribution $a^{t}$. 

At each decoding step, the previous context vector $c_{t-1}$ is concatenated with current input $x_t$, and fed through a non-linear recurrent function along with the previous hidden state $s_{t-1}$ to produce the new hidden state $s_t$.
The context vector $c_t$ is then calculated according to \eqnref{eq:att} and concatenated with the decoder state $s_t$ to produce the logits for the vocabulary distribution $P_{vocab}$ at decode step $t$:
 $P^t_{vocab} = \mathrm{softmax}(V_2(V_1[s_t,c_t] + b_1) + b_2)$, where $V_1$, $V_2$, $b_1$, $b_2$ are learnable parameters. 
To enable copying out-of-vocabulary words from source text, a pointer similar to \newcite{Vinyals:15} is built upon the attention distribution and controlled by the generation probability $p_{gen}$:
\begin{equation*} \label{eq:pg}
\begin{split}
p^t_{gen} & = \sigma(U_{c} c_t + U_{s} s_t + U_{x} x_t + b_{ptr}) \\
P_{attn}^t(w) & = p^t_{gen} P^t_{vocab}(w) + (1-p^t_{gen}) \sum_{i:w_i=w}a_{i}^t
\end{split}
\end{equation*}
where $U_{c}$, $U_s$, $U_x$, and $b_{ptr}$ are learnable parameters. $x_t$ and $s_t$ are the input token and decoder's state at $t^{th}$ decoding step. $\sigma$ is the sigmoid function. We can see $p_{gen}$ as a soft gate that controls the model's behavior of copying from text with attention distribution $a_{i}^t$ versus sampling from vocabulary with generation distribution $P^t_{vocab}$.

\subsection{Closed-Book Decoder}
As shown in \eqnref{eq:att}, the attention distribution $a_i$ depends on decoder's hidden state $s_t$, which is derived from decoder's memory state $c_t$. If $c_t$ does not encode salient information from the source text or encodes too much unimportant information, the decoder will have a hard time to locate relevant encoder states with attention.
However, as explained in the introduction, most gradients are back-propagated through attention layer to the encoder's hidden state $h_t$, not directly to the final memory state, and thus provide little incentive for the encoder to memorize salient information in $c_t$. 

Therefore, to enhance encoder's memory, we add a closed-book decoder, which is a unidirectional LSTM decoder 
without attention/pointer layer. The two decoders share a single encoder and word-embedding matrix, while out-of-vocabulary (OOV) words are simply represented as [UNK] for the closed-book decoder. The entire 2-decoder model is represented in \figref{fig:model}.
During training, we optimize the weighted sum of negative log likelihoods from the two decoders:
\vspace{-5pt}
\begin{equation} \label{eq:total-loss}
\begin{split}
\mathcal{L}_{XE} = \frac{1}{T} \sum_{t=1}^T &-((1-\gamma)\log P^t_{attn}(w|x_{1:t}) \\
&+ \gamma\log P^t_{cbdec}(w|x_{1:t}))
\end{split}
\vspace{-5pt}
\end{equation}
where $P_{cbdec}$ is the generation probability from the closed-book decoder. The mix ratio $\gamma$ is tuned on the validation set.

\subsection{Reinforcement Learning}
In the reinforcement learning setting, our summarization model is the policy network that generates words to form a summary.
Following~\newcite{Paulus:17}, we use a self-critical policy gradient training algorithm \cite{Rennie:16,Williams:92} for both our baseline and 2-decoder model. For each passage, we sample a summary $y^s = w^s_{1:T+1}$, and greedily generate a summary $\hat{y}= \hat{w}_{1:T+1}$ by selecting the word with the highest probability at each step.
Then these two summaries are fed to a reward function $r$, which is the ROUGE-L scores in our case. The RL loss function is:
\vspace{-3pt}
\begin{equation} \label{eq:rl-loss}
\mathcal{L}_{RL} = \frac{1}{T} \sum_{t=1}^T (r(\hat{y}) - r(y^s)) \log P^t_{attn}(w_{t+1}^s | w_{1:t}^s)
\vspace{-3pt}
\end{equation}
where the reward for the greedily-generated summary ($r(\hat{y})$) acts as a baseline to reduce variance.
We train our reinforced model using the mixture of \eqnref{eq:rl-loss} and \eqnref{eq:total-loss}, since \newcite{Paulus:17} showed that a pure RL objective would lead to summaries that receive high rewards but are not fluent. The final mixed loss function for RL is: $\mathcal{L}_{XE+RL} = \lambda\mathcal{L}_{RL} + (1-\lambda)\mathcal{L}_{XE}$, where the value of $\lambda$ is tuned on the validation set.

\begin{table}[t] 
\begin{small}
\centering
\begin{tabular}{|c | c |c |c | c |} 
\hline
\multirow{2}*{} & \multicolumn{3}{|c|}{ROUGE} & MTR\\\cline{2-5}
 & 1 & 2 & L & Full\\
\hline
\multicolumn{5}{|c|}{\textsc{previous works}} \\
\hline
$^\star$(Nallapati16) & 35.46 & 13.30 & 32.65 & \\
pg (See17) & 36.44 & 15.66 & 33.42 & 16.65 \\
\hline
\multicolumn{5}{|c|}{\textsc{our models}} \\
\hline
pg (baseline) & 36.70 & 15.71 & 33.74 & 16.94 \\
pg + cbdec & 38.21 & 16.45 & 34.70 & 18.37\\
\hline
RL + pg & 37.02 & 15.79 & 34.00 & 17.55 \\
RL + pg + cbdec & \textbf{38.58} & \textbf{16.57} & \textbf{35.03} & \textbf{18.86} \\
\hline
\end{tabular}
\caption{ROUGE F1 and METEOR scores (\emph{non-coverage}) on CNN/Daily Mail test set of previous works and our models. `pg' is the pointer-generator baseline, and `pg + cbdec' is our 2-decoder model with closed-book decoder(cbdec). The model marked with $\star$ is trained and evaluated on the anonymized version of the data.}
\label{table:non-coverage}
\end{small}
\end{table}

\section{Experimental Setup}
We evaluate our models mainly on \emph{CNN/Daily Mail} dataset \cite{Hermann:15,Nallapati:16}, which is a large-scale, long-paragraph summarization dataset. It has online news articles (781 tokens or \texttildelow 40 sentences on average) with paired human-generated summaries (56 tokens or 3.75 sentences on average). The entire dataset has 287,226
training pairs, 13,368 validation pairs and 11,490
test pairs. We use the same version of data as \newcite{See:17}, which is the original text with no preprocessing to replace named entities.
We also use DUC-2002, which is also a long-paragraph summarization dataset of news articles. This dataset has 567 articles and 1\texttildelow 2 summaries per article. 

All the training details (e.g., vocabulary size, RNN dimension, optimizer, batch size, learning rate, etc.) are provided in the supplementary materials.

\section{Results}

\begin{table}
\begin{small}
\centering
\begin{tabular}{|c | c |c |c | c |} 
\hline
\multirow{2}*{} & \multicolumn{3}{|c|}{ROUGE} & MTR\\\cline{2-5}
 & 1 & 2 & L & Full\\
\hline
\multicolumn{5}{|c|}{\textsc{previous works}} \\
\hline
pg (See17) & 39.53 & 17.28 & 36.38 & 18.72 \\
RL$^\star$ (Paulus17) & 39.87 & 15.82 & 36.90 & \\ 
\hline
\multicolumn{5}{|c|}{\textsc{our models}} \\
\hline
pg (baseline) & 39.22 & 17.02 & 35.95 & 18.70 \\
pg + cbdec & 40.05 & 17.66 & 36.73 & 19.48 \\
\hline
RL + pg & 39.59 & 17.18 & 36.16 & 19.70 \\
RL + pg + cbdec & \textbf{40.66} & \textbf{17.87} & \textbf{37.06} & \textbf{20.51} \\
\hline
\end{tabular}
\caption{ROUGE F1 and METEOR scores (\emph{with-coverage}) on the \emph{CNN/Daily Mail} test set. Coverage mechanism~\cite{See:17} is used in all models except the RL model~\cite{Paulus:17}.
The model marked with $\star$ is trained and evaluated on the anonymized version of the data.
}
\label{table:main}
\end{small}
\end{table}

\begin{table}[t] 
\begin{small}
\centering
\begin{tabular}{|c | c |c |c | c |} 
\hline
\multirow{2}*{} & \multicolumn{3}{|c|}{ROUGE} & MTR\\\cline{2-5}
 & 1 & 2 & L & Full\\
\hline
pg (See17) & 37.22 & 15.78 & 33.90 & 13.69 \\
pg (baseline) & 37.15 & 15.68 & 33.92 & 13.65 \\
pg + cbdec & 37.59 & 16.84 & 34.43 & 13.82 \\
\hline
RL + pg & 39.92 & 16.71 & 36.13 & 15.12 \\
RL + pg + cbdec & \textbf{41.48} & \textbf{18.69} & \textbf{37.71} & \textbf{15.88} \\
\hline
\end{tabular}
\caption{ROUGE F1 and METEOR scores on DUC-2002 (test-only transfer setup).\label{table:DUC-2002} \vspace{-1pt}}
\end{small}
\end{table}

\begin{figure}[t]
\centering
\includegraphics[width=0.43\textwidth]{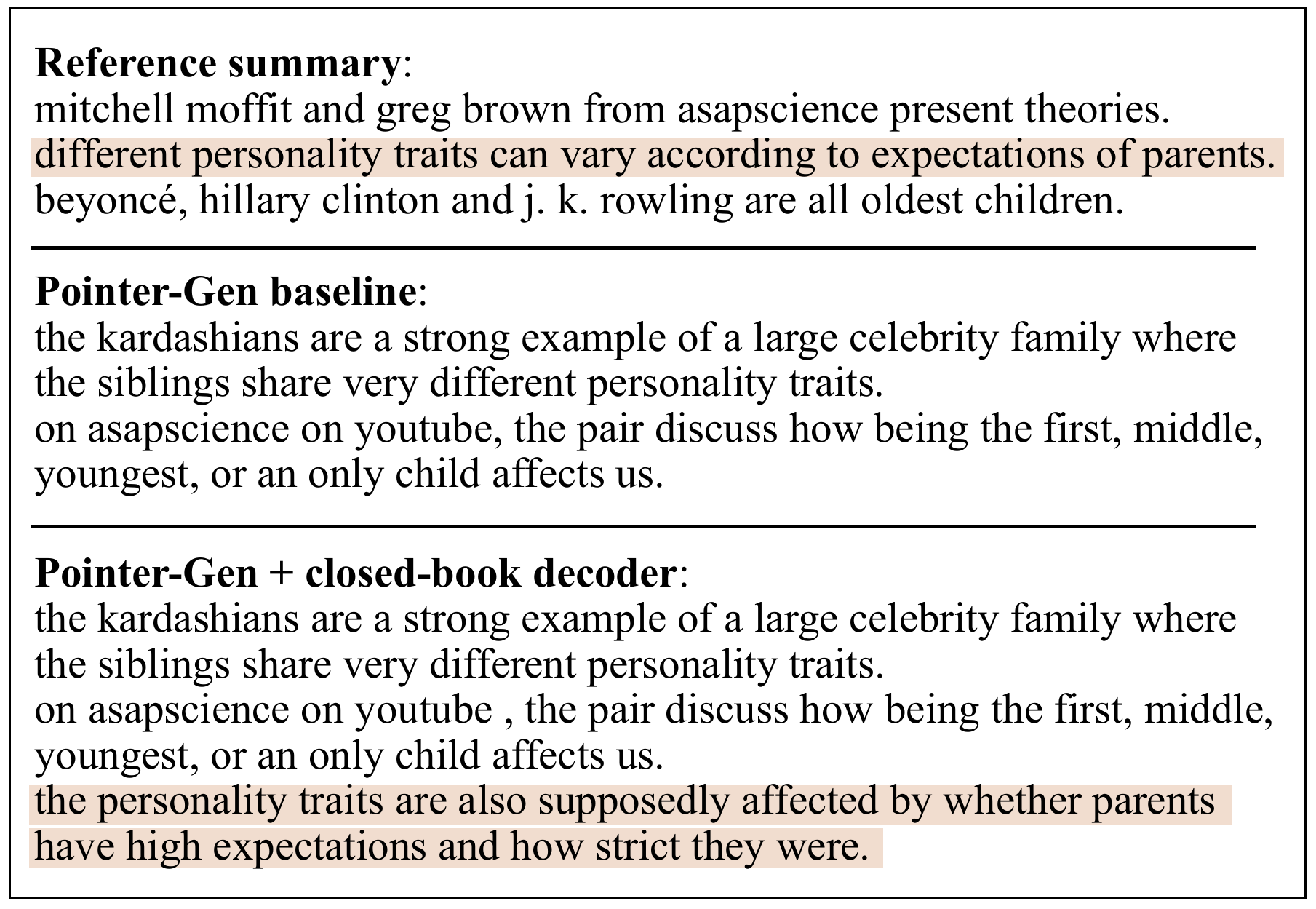}
\vspace{-8pt} 
\caption{The summary generated by our 2-decoder model covers salient information (highlighted in red) mentioned in the reference summary, which is not presented in the baseline summary.\label{fig:example-kardashians}}
\end{figure}

We first report our evaluation results on \emph{CNN/Daily Mail} dataset. 
As shown in \tabref{table:non-coverage}, our 2-decoder model achieves statistically significant improvements\footnote{Our improvements in Table~\ref{table:non-coverage} are statistically significant with $p<0.001$ (using bootstrapped randomization test with 100k samples~\cite{Efron:94}) and have a 95\% ROUGE-significance interval of at most \textpm 0.25.} upon the pointer-generator baseline (pg), with +1.51, +0.74, and +0.96 points advantage in ROUGE-1, ROUGE-2 and ROUGE-L~\cite{Lin:04}, and +1.43 points advantage in METEOR~\cite{Denkowski:14}. In the reinforced setting, our 2-decoder model still maintains significant ($p<0.001$) advantage in all metrics over the pointer-generator baseline.

We further add the coverage mechanism as in \newcite{See:17} to both baseline and 2-decoder model, and our 2-decoder model (pg + cbdec) again receives significantly higher\footnote{All our improvements in Table~\ref{table:main} are statistically significant with $p<0.001$, and have a 95\% ROUGE-significance interval of at most \textpm 0.25.} scores than the original pointer-generator (pg) from~\newcite{See:17} and our own pg baseline, in all ROUGE and METEOR metrics (see \tabref{table:main}).
In the reinforced setting, our 2-decoder model (RL + pg + cbdec) outperforms our strong RL baseline (RL + pg) by a considerable margin (stat. significance of $p<0.001$). \figref{fig:example-infant-body-small} and \figref{fig:example-kardashians} show two examples of our 2-decoder model generating summaries that cover more salient information than those generated by the pointer-generator baseline (see supplementary materials for more example summaries).

We also evaluate our 2-decoder model with coverage on the DUC-2002 test-only generalizability/transfer setup by decoding the entire dataset with our models pre-trained on \emph{CNN/Daily Mail}, again achieving decent improvements (shown in \tabref{table:DUC-2002}) over the single-decoder baseline as well as~\newcite{See:17}, in both a cross-entropy and a reinforcement learning setup.

\begin{table}[t] 
\begin{small}
\centering
\begin{tabular}{|l|c|} 
\hline
Model & Score \\
\hline
2-Decoder Wins & 49 \\
Pointer-Generator Wins & 31 \\
Non-distinguishable & 20 \\
\hline
\end{tabular}
\vspace{-3pt}
\caption{Human evaluation for our 2-decoder model versus the pointer-generator baseline.
\label{table:human-eval}
\vspace{-1pt}
}
\end{small}
\end{table}

\subsection{Human Evaluation}
We also conducted a small-scale human evaluation study by randomly selecting 100 samples from the \emph{CNN/DM} test set and then asking human annotators to rank the baseline summaries versus the 2-decoder's summaries (randomly shuffled to anonymize model identity) according to an overall score based on readability (grammar, fluency, coherence) and relevance (saliency, redundancy, correctness). 
As shown in \tabref{table:human-eval}, our 2-decoder model outperforms the pointer-generator baseline (stat. significance of $p<0.03$).

\section{Analysis}
In this section, we present a series of analysis and tests in order to understand the improvements of the 2-decoder models reported in the previous section, and to prove that it fulfills our intuition that the closed-book decoder improves the encoder's ability to encode salient information in the memory state.

\begin{table}[t]
\begin{small}
\centering
\begin{tabular}{|c|c|c|} 
\hline
& similarity \\[1.5pt]
\hline
pg (baseline) &0.817 \\[1.5pt]
pg + cbdec ($\gamma=\frac{1}{2}$)& 0.869 \\[1.5pt]
pg + cbdec ($\gamma=\frac{2}{3}$)& \textbf{0.889} \\[1.5pt]
pg + cbdec ($\gamma=\frac{5}{6}$)& 0.872 \\[1.5pt]
pg + cbdec ($\gamma=\frac{10}{11}$)& 0.860 \\[1.5pt]
\hline
\end{tabular}
\vspace{-3pt}
\caption{Cosine-similarity between memory states after two forward passes.}
\label{table:similarity}
\vspace{-5pt}
\end{small}
\end{table}

\subsection{Memory Similarity Test}

To verify our argument that the closed-book decoder improves the encoder's memory ability, we design a test to numerically evaluate the representation power of encoder's final memory state. 
We perform two forward passes for each encoder (2-decoder versus pointer-generator baseline). For the first pass, we feed the entire article to the encoder and collect the final memory state; for the second pass we feed the ground-truth summary to the encoder and collect the final memory state. Then we calculate the cosine similarity between these two memory-state vectors. 
For an optimal summarization model, its encoder's memory state after reading the entire article should be highly similar to its memory state after reading the ground truth summary (which contains all the important information), because this shows that when reading a long passage, the model is only encoding important information in its memory and forgets the unimportant information. 
The results in \tabref{table:similarity} show that the encoder of our 2-decoder model achieves significantly ($p<0.001$) higher article-summary similarity score than the encoder of a pointer-generator baseline. This observation verifies our hypothesis that the closed-book decoder can improve the memory ability of the encoder.

\subsection{Ablation Studies and Sanity Check}

\noindent\textbf{Fixed-Encoder Ablation}:
Next, we conduct an ablation study in order to prove the qualitative superiority of our 2-decoder model's encoder to the baseline encoder. To do this, we train two pointer-generators with randomly initialized decoders and word embeddings. For the first model, we restore the pre-trained encoder from our pointer-generator baseline; for the second model, we restore the pre-trained encoder from our 2-decoder model. We then fix the encoder's parameters for both models during the training, only updating the embeddings and decoders with gradient descent.
As shown in the upper half of \tabref{table:ablation}, the pointer-generator with our 2-decoder model's encoder receives significantly higher ($p<0.001$) scores in ROUGE than the pointer-generator with baseline's encoder. Since these two models have the exact same structure with only the encoders initialized according to different pre-trained models, the significant improvements in metric scores suggest that our 2-decoder model does have a stronger encoder than the pointer-generator baseline.

\begin{figure}[t]
\centering
\includegraphics[width=0.35\textwidth]{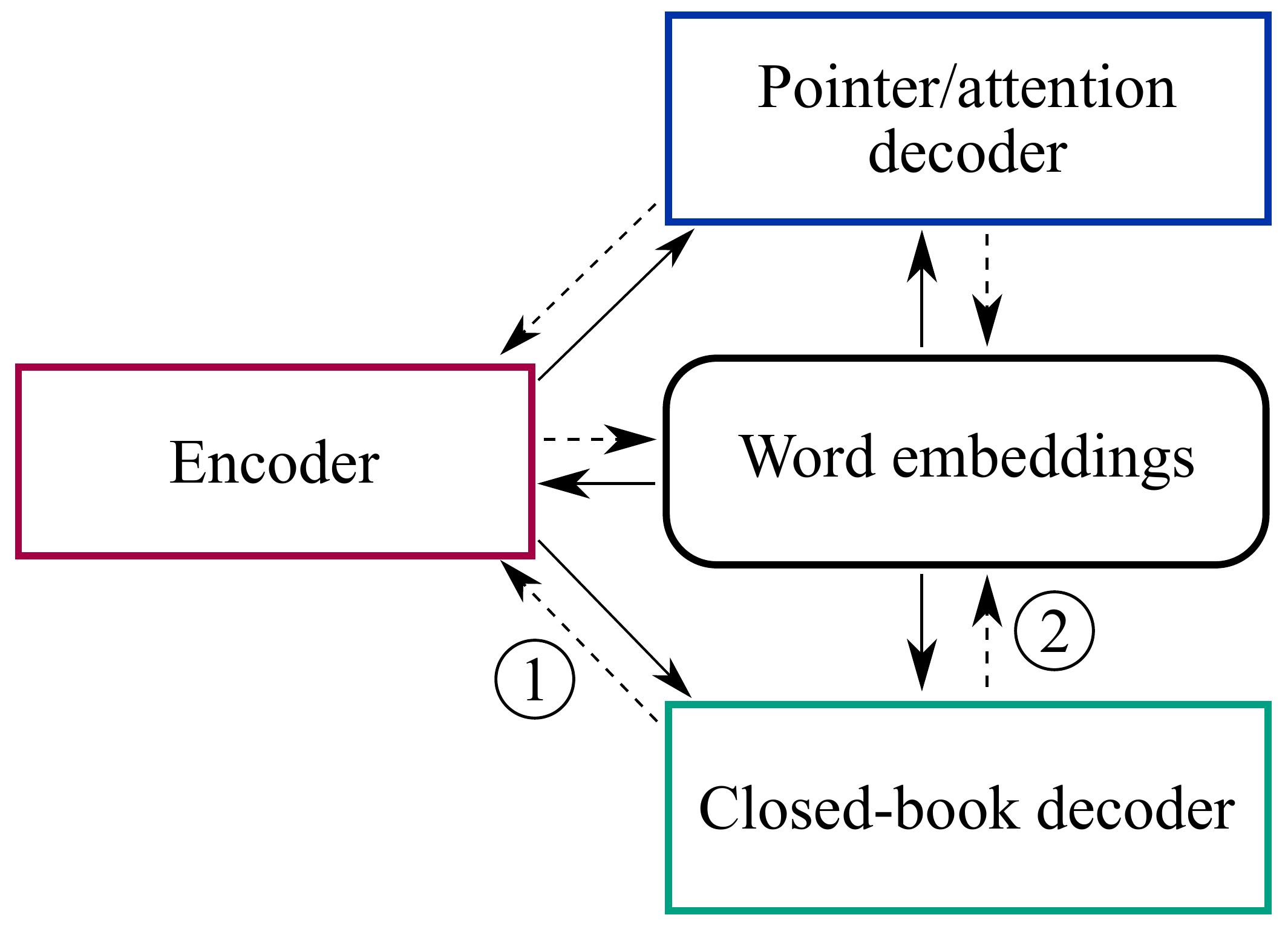}
\vspace{-10pt} 
\caption{Solid lines represent the forward pass, and dashed lines represent the gradient flow in back-propagation.
For the two ablation tests, we stop the gradient at \raisebox{.5pt}{\textcircled{\raisebox{-.9pt} {1}}} and \raisebox{.5pt}{\textcircled{\raisebox{-.9pt} {2}}} respectively.}
\label{fig:gradient-flow-cut}
\vspace{-5pt}
\end{figure}

\noindent\textbf{Gradient-Flow-Cut Ablation}:
We further design another ablation test to identify how the gradients from the closed-book decoder influence the entire model during training. \figref{fig:gradient-flow-cut} demonstrates the forward pass (solid line) and gradient flow (dashed line) between encoder, decoders, and embeddings in our 2-decoder model. As we can see, the closed-book decoder only depends on the word embeddings and encoder. Therefore it can affect the entire model during training by influencing either the encoder or the word-embedding matrix. 
When we stop the gradient flow between the encoder and closed-book decoder (\raisebox{.5pt}{\textcircled{\raisebox{-.9pt} {1}}} in \figref{fig:gradient-flow-cut}), and keep the flow between closed-book decoder and embedding matrix (\raisebox{.5pt}{\textcircled{\raisebox{-.9pt} {2}}} in \figref{fig:gradient-flow-cut}), we observe non-significant improvements in ROUGE compared to the baseline. On the other hand, when we stop the gradient flow at \raisebox{.5pt}{\textcircled{\raisebox{-.9pt} {2}}} and keep \raisebox{.5pt}{\textcircled{\raisebox{-.9pt} {1}}}, the improvements are statistically significant ($p<0.01$) (see the lower half of \tabref{table:ablation}). This proves that the gradients back-propagated from closed-book decoder to the encoder can strengthen the entire model, and hence verifies the gradient-flow intuition discussed in introduction (Sec.~\ref{sec 1}).

\begin{table}[t] 
\begin{small}
\centering
\begin{tabular}{|c | c |c |c|} 
\hline
\multirow{2}*{} & \multicolumn{3}{|c|}{ROUGE} \\ \cline{2-4}
 &1&2&L\\
\hline
\multicolumn{4}{|c|}{\textsc{Fixed-encoder ablation}} \\
\hline
pg baseline's encoder & 37.59 & 16.27 & 34.33 \\
2-decoder's encoder & \textbf{38.44} & \textbf{16.85} & \textbf{35.17} \\
\hline
\multicolumn{4}{|c|}{\textsc{Gradient-Flow-Cut ablation}} \\
\hline
pg baseline & 37.73 & 16.52 & 34.49 \\
stop \raisebox{.5pt}{\textcircled{\raisebox{-.9pt} {1}}} & 37.72 & 16.58 & 34.54 \\
stop \raisebox{.5pt}{\textcircled{\raisebox{-.9pt} {2}}} & \textbf{38.35} & \textbf{16.79} & \textbf{35.13} \\
\hline
\end{tabular}
\vspace{-5pt}
\caption{ROUGE F1 scores of ablation studies, evaluated on \emph{CNN/Daily Mail} validation set.}
\label{table:ablation}
\vspace{-2pt}
\end{small}
\end{table}

\begin{table}[t] 
\begin{small}
\centering
\begin{tabular}{|c | c |c |c|} 
\hline
\multirow{2}*{} & \multicolumn{3}{|c|}{ROUGE} \\ \cline{2-4}
 &1&2&L\\
\hline
pg baseline & 37.73 & 16.52 & 34.49 \\
pg + ptrdec & 37.66 & 16.50 & 34.47\\
pg-2layer & 37.92 & 16.48 & 34.62\\
pg-big & 38.03 & 16.71 & 34.84\\
pg + cbdec & \textbf{38.87} & \textbf{16.93} & \textbf{35.38}\\
\hline
\end{tabular}
\caption{ROUGE F1 and METEOR scores of sanity check ablations, evaluated on \emph{CNN/DM} validation set. 
\label{table:sanity-check} \vspace{0pt}
}
\end{small}
\end{table}

\noindent\textbf{Model Capacity}:
\label{model_capacity}
To validate and sanity-check that the improvements are the result of the inclusion of our closed-book decoder and not due to some trivial effects of having two decoders or larger model capacity (more parameters), we train a variant of our model with two duplicated (initialized to be different) attention-pointer decoders. 
We also evaluate a pointer-generator baseline with 2-layer encoder and decoder (pg-2layer) and increase the LSTM hidden dimension and word embedding dimension of the pointer-generator baseline (pg-big) to exceed the total number of parameters of our 2-decoder model (34.5M versus 34.4M parameters).
\tabref{table:sanity-check} shows that neither of these variants can match our 2-decoder model in terms of ROUGE and METEOR scores, and hence proves that the improvements of our model are indeed because of the closed-book decoder rather than due to simply having more parameters.\footnote{It is also important to point out that our model is not a 2-decoder ensemble, because we use only the pointer decoder during inference. Therefore, the number of parameters used for inference is the same as the pointer-generator baseline.}

\begin{table}[t] 
\begin{small}
\centering
\begin{tabular}{|c | c |c |c |} 
\hline
\multirow{2}*{} & \multicolumn{3}{|c|}{ROUGE}\\\cline{2-4}
 & 1 & 2 & L \\
\hline
$\gamma=0$ & 37.73 & 16.52 & 34.49 \\
$\gamma=1/2$ & 38.09 & 16.71 & 34.89 \\
$\gamma=2/3$ & \textbf{38.87} & \textbf{16.93} & \textbf{35.38} \\
$\gamma=5/6$ & 38.21 & 16.69 & 34.81\\
$\gamma=10/11$ & 37.99 & 16.39 & 34.7\\
\hline
\end{tabular}
\caption{ROUGE F1 scores on \emph{CNN/DM} validation set, of 2-decoder models with different values of the closed-book-decoder:pointer-decoder mixed loss ratio.}
\label{table:validation}
\vspace{-5pt}
\end{small}
\end{table}
\noindent\textbf{Mixed-loss Ratio Ablation}:
We also present evaluation results (on the \emph{CNN/Daily Mail} validation set) of our 2-decoder models with different closed-book-decoder:pointer-decoder mixed-loss ratio ($\gamma$ in \eqnref{eq:total-loss}) in \tabref{table:validation}. 
The model achieves the best ROUGE and METEOR scores at $\gamma=\frac{2}{3}$. Comparing \tabref{table:validation} with \tabref{table:similarity}, we observe a similar trend between the increasing ROUGE/METEOR scores and increasing memory cosine-similarities, which suggests that the performance of a pointer-generator is strongly correlated with the representation power of the encoder's final memory state.

\begin{table}[t] 
\begin{small}
\centering
\begin{tabular}{| c | c| c|} 
\hline
& saliency 1 & saliency 2 \\
\hline
pg (See17) & 60.4\% & 27.95\%\\
our pg baseline & 59.6\% & 28.95\%\\
pg + cbdec & 62.1\% & 29.97\%\\
RL + pg & 62.5\% & 30.17\%\\
RL + pg + cbdec & \textbf{66.2\%} & \textbf{31.40\%}\\
\hline
\end{tabular}
\vspace{-3pt}
\caption{Saliency scores based on \emph{CNN/Daily Mail} cloze blank-filling task and a keyword-detection approach~\cite{Ram18summ}. All models in this table are trained with coverage loss.}
\label{table:saliency}
\end{small}
\end{table}

\subsection{Saliency and Repetition}
Finally, we show that our 2-decoder model can make use of this better encoder memory state to summarize more salient information from the source text, as well as to avoid generating unnecessarily lengthy and repeated sentences besides achieving significant improvements on ROUGE and METEOR metrics.

\begin{table}[t] 
\begin{small}
\centering
\begin{tabular}{|c|c|c|c|c|} 
\hline
& 3-gram & 4-gram & 5-gram & sent \\
\hline
pg (baseline) & 13.20\% & 12.32\% & 11.60\% &8.39\%\\
pg + cbdec & \textbf{9.66\%} & \textbf{9.02\%} & \textbf{8.55\%} &\textbf{6.72\%}\\
\hline
\end{tabular}
\vspace{-3pt}
\caption{Percentage of repeated 3, 4, 5-grams and sentences in generated summaries.}
\label{table:repetition}
\vspace{-5pt}
\end{small}
\end{table}

\noindent\textbf{Saliency}:
To evaluate saliency, we design a keyword-matching test based on the original \emph{CNN/Daily Mail} cloze blank-filling task~\cite{Hermann:15}. Each news article in the dataset is marked with a few cloze-blank keywords that represent salient entities, including names, locations, etc. 
We count the number of keywords that appear in our generated summaries, and found that the output of our best teacher-forcing model (pg+cbdec with coverage) contains 62.1\% of those keywords, while the output provided by \newcite{See:17} has only 60.4\% covered. Our reinforced 2-decoder model (RL + pg + cbdec) further increases this percentage to $66.2\%$.
The full comparison is shown in the first column of \tabref{table:saliency}. 
We also use the saliency metric in \newcite{Ram18summ}, which finds important words detected via a keyword classifier (trained on the SQuAD dataset~\cite{rajpurkar2016squad}). The results are shown in the second column of \tabref{table:saliency}.
Both saliency tests again demonstrate our 2-decoder model's ability to memorize important information and address them properly in the generated summary. \figref{fig:example-infant-body-small} and \figref{fig:example-kardashians} show two examples of summaries generated by our 2-decoder model compared to baseline summaries.

\noindent\textbf{Summary Length}:
On average, summaries generated by our 2-decoder model have 66.42 words per summary, while the pointer-generator-baseline summaries have 65.88 words per summary (and the same effect holds true for RL models, where there is less than 1-word difference in average length). This shows that our 2-decoder model is able to achieve higher saliency with similar-length summaries (i.e., it is not capturing more salient content simply by generating longer summaries).

\noindent\textbf{Repetition}:
We observe that out 2-decoder model can generate summaries that are less redundant compared to the baseline, when both models are not trained with coverage mechanism. \tabref{table:repetition} shows the percentage of repeated n-grams/sentences in summaries generated by the pointer-generator baseline and our 2-decoder model. 

\noindent\textbf{Abstractiveness}:
Abstractiveness is another major challenge for current abstractive summarization models other than saliency. 
Since our baseline is an abstractive model, we measure the percentage of novel n-grams (n=2, 3, 4) in our generated summaries, and find that our 2-decoder model generates 1.8\%, 4.8\%, 7.6\% novel n-grams while our baseline summaries have 1.6\%, 4.4\%, 7.1\% on the same test set. Even though generating more abstractive summaries is not our focus in this paper, we still show that our improvements in metric and saliency scores are not obtained at the cost of making the model more extractive.

\subsection{Discussion: Connection to Multi-Task Learning}
Our 2-decoder model somewhat resembles a Multi-Task Learning (MTL) model, in that both try to improve the model with extra knowledge that is not available to the original single-task baseline. 
While our model uses MTL-style parameter sharing to introduce extra knowledge from the same dataset, traditional Multi-Task Learning usually employs additional/out-of-domain auxiliary tasks/datasets as related knowledge (e.g., translation with 2 language-pairs). Our 2-decoder model is more about how to learn to do a single task from two different points of view, as the pointer decoder is a hybrid of extractive and abstractive summarization models (primary view), and the closed-book decoder is trained for abstractive summarization only (auxiliary view). The two decoders share their encoder and embeddings, which helps enrich the encoder's final memory state representation.

Moreover, as shown in Sec.~\ref{model_capacity}, our 2-decoder model (pg + cbdec) significantly outperforms the 2-duplicate-decoder model (pg + ptrdec) as well as single-decoder models with more layers/parameters, hence proving that our design of the auxiliary view (closed-book decoder doing abstractive summarization) is the reason behind the improved performance, rather than some simplistic effects of having a 2-decoder ensemble or higher \#parameters.

\section{Conclusion}
We presented a 2-decoder sequence-to-sequence architecture for summarization with a closed-book decoder that helps the encoder to better memorize salient information from the source text. On \emph{CNN/Daily Mail} dataset, our proposed model significantly outperforms the pointer-generator baselines in terms of ROUGE and METEOR scores (in both a cross-entropy (XE) setup and a reinforcement learning (RL) setup).
It also achieves improvements in a test-only transfer setup on the DUC-2002 dataset in both XE and RL cases. We further showed that our 2-decoder model indeed has a stronger encoder with better memory capabilities, and can generate summaries with more salient information from the source text.
To the best of our knowledge, this is the first work that studies the ``representation power'' of the encoder’s final state in an encoder-decoder model. 
Furthermore, our simple, insightful 2-decoder architecture can also be useful for other tasks that require long-term memory from the encoder, e.g., long-context QA/dialogue and captioning for long videos.

\section{Acknowledgement}
We thank the reviewers for their helpful comments. This work was supported by DARPA (YFA17-D17AP00022), Google Faculty Research
Award, Bloomberg Data Science Research Grant, Nvidia GPU awards, and Amazon AWS. The views contained in
this article are those of the authors and not of the funding agency.

\bibliography{emnlp2018}
\bibliographystyle{acl_natbib_nourl}

\appendix
\noindent{\large\textbf{Supplementary Material}}
\label{sec:supplemental}

\section{Coverage Mechanism}

\newcite{See:17} apply coverage mechanism to the pointer-generator in order to alleviate repetition. They maintain a coverage vector $c^t$ as the sum of attention distribution over all previous decoding steps $1:t-1$. This vector is incorporated in calculating the attention distribution at current step $t$:
\begin{equation} \label{eq:coverage}
\begin{split}
c^t & = \sum_{t'=0}^{t-1} a^{t'} \\
e_{i}^{t} & = v^T \mathrm{tanh}(W_h h_i + W_s s_t +W_c c_i^t + b_{attn})
\end{split}
\end{equation}
where $W_h, W_s, W_c, b_{attn}$ are learnable parameters. They define the coverage loss and combine it with the primary loss to form a new loss function, which is used to fine-tune a converged pointer-generator model.
\begin{equation} \label{eq:cov-loss}
\begin{split}
\mathrm{loss}^{t}_{cov} &= \sum_{i} \mathrm{min} (a_i^t,c_i^t)\\
\mathcal{L}_{total} &= \frac{1}{T} \sum_{t=1}^T (-\log P^t_{pg}(w | x_{1:t}) + \bm{\lambda} \mathrm{loss}_{cov}^t)
\end{split}
\end{equation}

\section{Reinforcement Learning}
To overcome the exposure bias~\cite{Bengio:15} between training and testing, previous works \cite{Ranzato:16,Paulus:17} use reinforcement learning algorithms to directly optimize on metric scores for summarization models. In this setting, the generation of discrete words in a sentence is a sequence of \emph{actions}. The decision to take what \emph{action} is based on a \emph{Policy Network} $\pi_{\theta}$, which outputs a distribution of all possible \emph{actions} at that step. In our case, $\pi_{\theta}$ is simply our summarization model.

The process of generating a summary $\mathbf{s}$ given the source passage $\mathbf{P}$ can be summarized as follows. At each time step $t$, we sample a discrete \emph{action} $w_t \in \mathcal{V}$ - word in vocabulary, based on distribution from \emph{policy} $\pi_{\theta}(\mathbf{P}, \mathbf{s}_t)$, where $\mathbf{s}_t=w_{1:t-1}$ is the sequence of \emph{actions} sampled in previous steps. When we reach the end of the sequence at terminal step $T$ (end-of-sentence marker is sampled from $\pi_{\theta}$), we feed the entire sequence $\mathbf{s}_T = w_{1:T}$ into a reward function and get a reward $R(w_{1:T}|\mathbf{P})$.

In typical Reinforcement Learning, an agent with \emph{policy} receives rewards at intermediate steps while the discount factor is used to balance long-term and short-term rewards. In our task, there is no intermediate rewards, only a final reward at terminal step $T$. Therefore, the value function of a partial sequence $\mathbf{c}_{t} = w_{1:t}$ is the expected reward at the terminal step.
\begin{equation} \label{eq:value}
V(w_{1:t} | \mathbf{P}) = \E_{w_{t+1:T}}[R(w_{1:t};w_{t+1:T} | \mathbf{P})]
\end{equation}

The objective of policy gradient is to maximize the average value starting from the initial state:
\begin{equation} \label{eq:avg-value}
J(\theta)=\frac{1}{N} \sum_{n=1}^{N}V(w_{0} | \mathbf{I})
\end{equation}
where N is the total number of examples in training set. The gradient of $V(w_{0} | \mathbf{P})$ is computed as below \cite{Williams:92}:

\begin{equation} \label{eq:pg-1}
\begin{split}
\E_{w_{2:T}}[\sum_{t=1}^{T} \sum_{w_{t} \in \mathcal{V}} \nabla_{\theta} & \pi_{\theta}(w_{t+1} | w_{1:t}, \mathbf{P}) \\
&\times Q(w_{1:t}, w_{t+1} | \mathbf{P})]
\end{split}
\end{equation}
where $Q(w_{1:t}, w_t | \mathbf{P})$ is the \emph{state-action} value for a particular \emph{action} $w_{t+1}$ at \emph{state} $w_{1:t}$ given source passage $\mathbf{P}$, and should be calculated as follow:
\begin{equation} \label{eq:state-action}
Q(w_{1:t}, w_{t+1} | \mathbf{P}) = \E_{w_{t+2:T}}[R(w_{1:t+1}; w_{t+2:T} | \mathbf{P})]
\end{equation}

Previous work \cite{Liu:16} adopts Monte Carlo Rollout to approximate this expectation. Here we simply use the terminal reward $R(w_{1:T} | \mathbf{P})$ as an estimation with large variance.
To compensate for the variance, we use a baseline estimator that doesn't change the validity of gradients \cite{Williams:92}. We further follow \newcite{Paulus:17} to use the self-critical policy gradient training algorithm \cite{Rennie:16,Williams:92}. For each iteration, we sample a summary $y^s = w^s_{1:T+1}$, and greedily generate a summary $\hat{y}= \hat{w}_{1:T+1}$ by selecting the word with the highest probability at each step.
Then these two summaries are fed to a reward function $r$ that evaluates their closeness to the ground-truth. We choose ROUGE-L scores as the reward function $r$ as in previous work \cite{Paulus:17}. The RL loss function is as follows:
\begin{equation} \label{eq:rl-loss-supp}
\mathcal{L}_{RL} = \frac{1}{T} \sum_{t=1}^T (r(\hat{y}) - r(y^s)) \log \pi_{\theta}(w_{t+1}^s | w_{1:t}^s)
\end{equation}
where the reward for the greedily-generated summary ($r(\hat{y})$) acts as a baseline to reduce variance.

\section{Training Details}
We keep most of hyper-parameters and settings the same as in \newcite{See:17}.
We use a bi-directional LSTM of 400 steps for the encoder, and a uni-directional LSTM of 100 steps for both decoders.
All of our encoder and decoder LSTMs have hidden dimension of 256, and the word embedding dimension is set to 128. Our pre-set vocabulary has a total of 50k word tokens including special tokens for start, end, and out-of-vocabulary(OOV) signals. The embedding matrix is learned from scratch and shared between the encoder and two decoders.

All of our teacher forcing models reported are trained with Adagrad \cite{Adagrad:11} with learning rate of 0.15 and an initial accumulator value of 0.1. The gradients are clipped to a maximum norm of 2.0. 
The batch size is set to 16. Our model with closed-book decoder converged in about 200,000 to 240,000 iterations and achieved the best result on the validation set in another 2k\texttildelow 3k iterations with coverage loss added.
We restore the best checkpoints (pre-coverage and post-coverage) and apply policy gradient (RL). 
For this phase of training, we choose Adam optimizer \cite{Adam:14} because of its time efficiency, and the learning rate is set to 0.000001. The RL-XE mixed-loss ratio ($\lambda$) is set to 0.9984.

\section{Examples}

\begin{figure*}[t]
\centering
\includegraphics[width=0.9\textwidth]{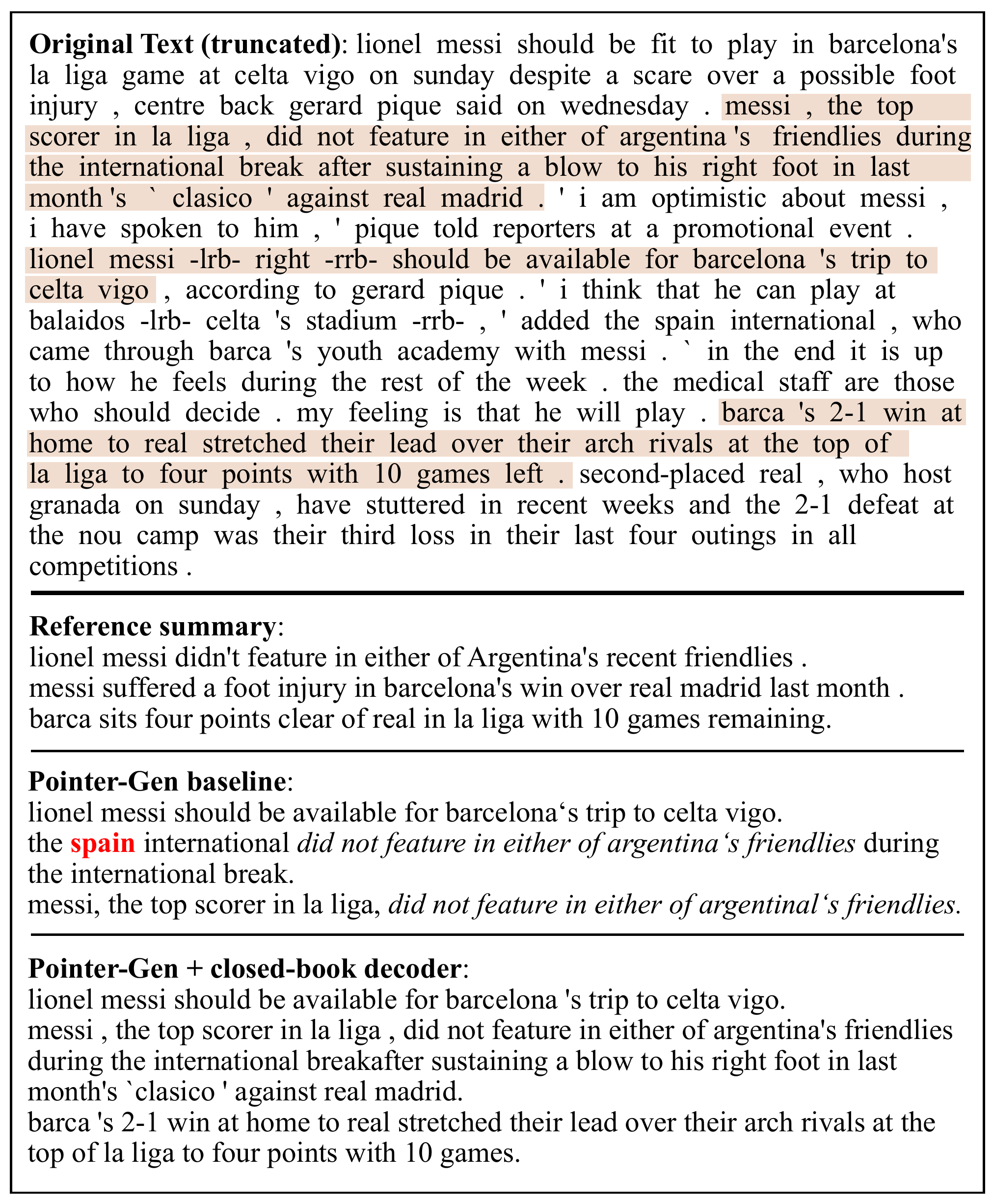}
\vspace{-10pt} 
\caption{The pointer-generator repeats itself (italic) and makes a factual error (red), while the 2-decoder (pointer-generator + closed-book decoder) generates the summary that recovers the salient information (highlighted) in the original text.}
\label{fig:example-messi}
\end{figure*}

\begin{figure*}[t]
\centering
\includegraphics[width=0.9\textwidth]{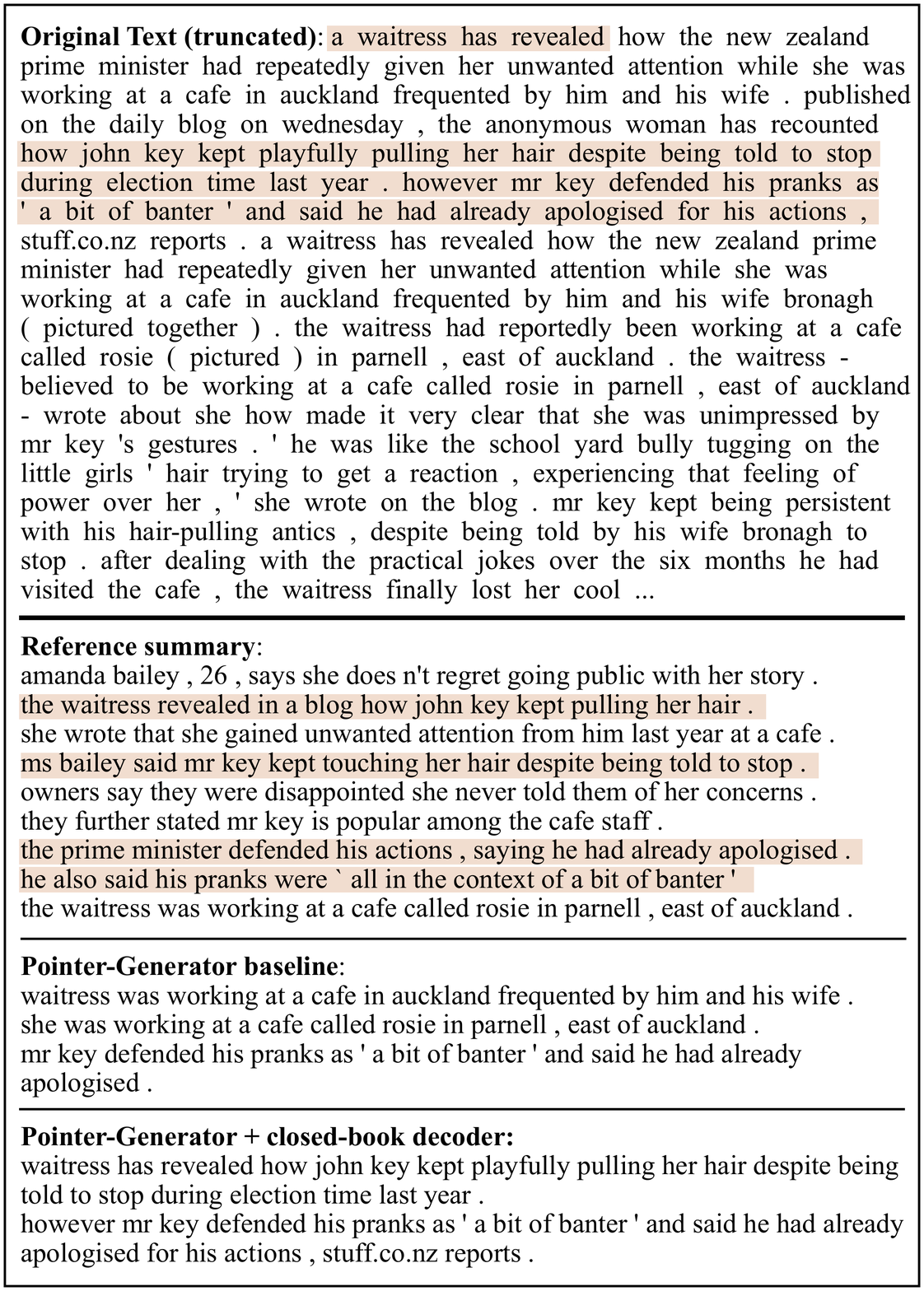}
\vspace{-10pt} 
\caption{The pointer-generator fails to address the most salient information from the original text, only mentioned a few unimportant points (where the waitress works), while the 2-decoder (pointer-generator + closed-book decoder) generates the summary that recovers the salient information (highlighted) in the original text.}
\label{fig:example-waitress}
\end{figure*}

\begin{figure*}[t]
\centering
\includegraphics[width=0.9\textwidth]{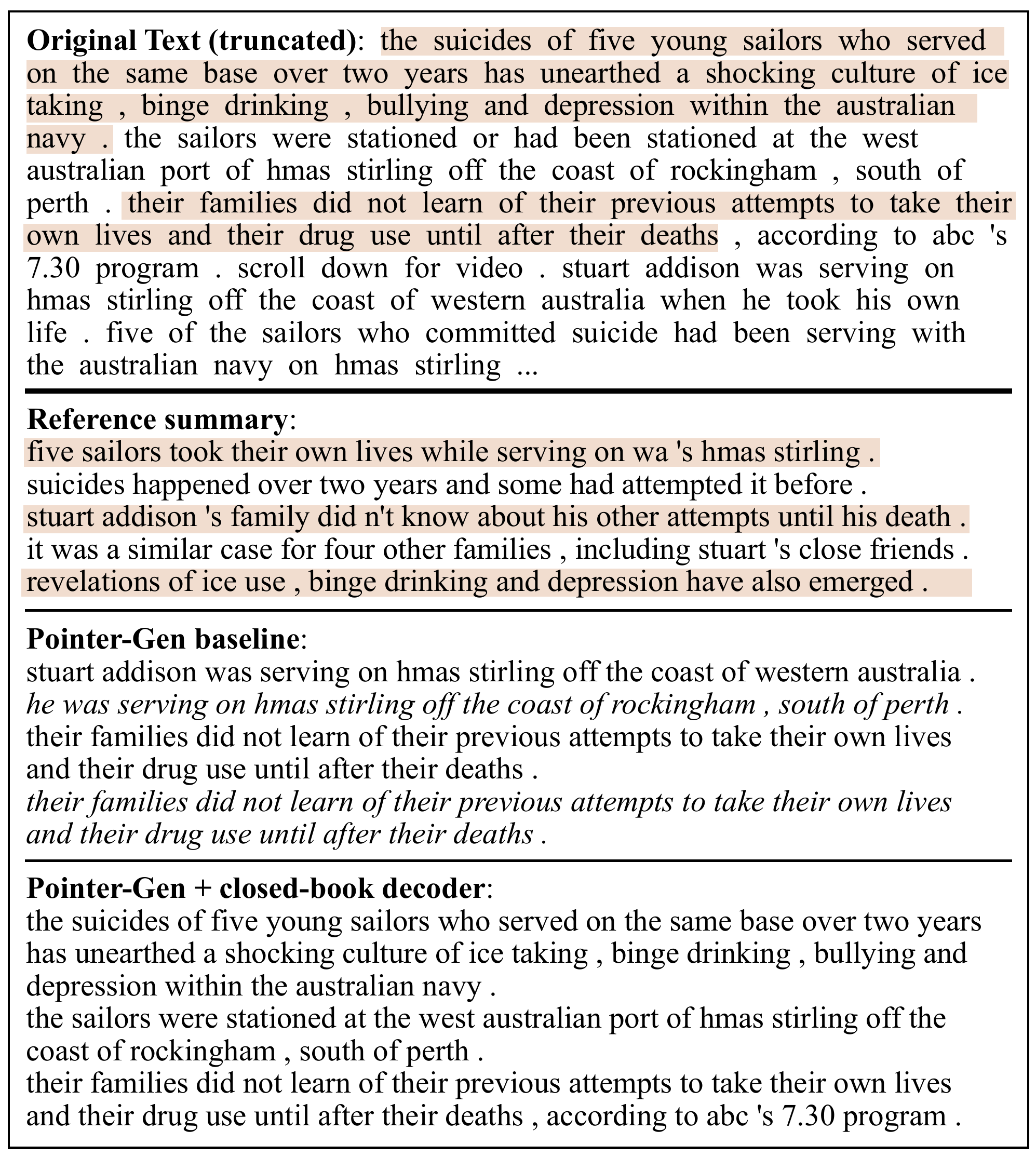}
\vspace{-10pt} 
\caption{The pointer-generator (non-coverage) repeats itself (italic), while the 2-decoder (pointer-generator + closed-book decoder) generates the summary that recovers the salient information (highlighted) in the original text as well as the reference summary.}
\label{fig:example-sailors}
\end{figure*}

We provide more example summaries generated by our 2-decoder and pointer-generator baseline (see \figref{fig:example-messi}, \figref{fig:example-waitress}, and \figref{fig:example-sailors} on the next page).

\end{document}